\documentclass[journal,twoside]{IEEEtran}
\IEEEoverridecommandlockouts

\usepackage{graphicx} \graphicspath{ {figures/} }
\usepackage{amsmath,amssymb,mathabx,amsbsy,mathtools,etoolbox}
\usepackage[utf8]{inputenc} 
\usepackage[T1]{fontenc}    
\usepackage{hyperref}
\usepackage{algorithmic}
\usepackage[ruled]{algorithm2e}

\usepackage{acronym}
\usepackage{enumitem}
\usepackage{balance}
\usepackage{xspace}
\usepackage{setspace}
\usepackage[skip=3pt,font=small]{subcaption}
\usepackage[skip=3pt,font=small]{caption}
\usepackage[dvipsnames,svgnames,x11names]{xcolor}
\usepackage[capitalise,nameinlink]{cleveref}
\usepackage{booktabs,tabularx,colortbl,multirow,multicol,array,makecell}
\usepackage{pifont}
\usepackage{cite}
\usepackage{nicematrix}

\makeatletter
\DeclareRobustCommand\onedot{\futurelet\@let@token\@onedot}
\def\@onedot{\ifx\@let@token.\else.\null\fi\xspace}

\def\etal{\emph{et al}\onedot}
\makeatother

\makeatletter
\def\BState{\State\hskip-\ALG@thistlm}
\makeatother

\makeatletter
\renewcommand{\paragraph}{%
  \@startsection{paragraph}{4}%
  {\z@}{0ex \@plus 0ex \@minus 0ex}{-1em}%
  {\hskip\parindent\normalfont\normalsize\bfseries}%
}
\makeatother

\crefname{algorithm}{Alg.}{Algs.}
\Crefname{algocf}{Algorithm}{Algorithms}
\crefname{section}{Sec.}{Secs.}
\Crefname{section}{Section}{Sections}
\crefname{table}{Tab.}{Tabs.}
\Crefname{table}{Table}{Tables}
\crefname{figure}{Fig.}{Fig.}
\Crefname{figure}{Figure}{Figure}

\definecolor{gblue}{HTML}{4285F4}
\definecolor{gred}{HTML}{DB4437}
\definecolor{ggreen}{HTML}{0F9D58}

\definecolor{mygray}{gray}{.92}

\acrodef{qp}[QP]{Quadratic Programming}
\acrodef{dof}[DoF]{Degree of Freedom}
\acrodef{ros}[ROS]{Robot Operating System}
\acrodef{dof}[DoF]{Degree of Freedom}
\acrodef{com}[CoM]{center of mass}
\acrodef{lipm}[LIPM]{Linear Inverted Pendulum Model}
\acrodef{ocp}[OCP]{Optimal Control Problem}
\acrodef{mpc}[MPC]{Model Predictive Control}
\acrodef{wbc}[WBC]{Whole-Body Control}
\acrodef{cmm}[CMM]{ Centroidal Momentum Matrix}
\acrodef{ik}[IK]{Inverse Kinematics}
\acrodef{rl}[RL]{Reinforcement Learning}
\acrodef{il}[IL]{Imitation Learning}
\acrodef{to}[TO]{Trajectory Optimization}
\acrodef{imu}[IMU]{Inertial Measurement Unit}
\acrodef{zmp}[ZMP]{Zero Moment Point}
\acrodef{mocap}[MoCap]{motion capture}

\title{Dynamic Whole-Body Dancing with Humanoid Robots --- A Model-Based Control Approach}
\author{Shibowen Zhang, Jiayang Wu, Guannan Liu, Helin Zhu, Junjie Liu, Zhehan Li, Junhong Guo, Xiaokun Leng, Hangxin Liu, Jingwen Zhang, Jikai Wang, Zonghai Chen, Zhicheng He, Jiayi Wang, Yao Su
\thanks{This work was supported in part by the National Natural Science Foundation of China (No. 62403063, 62403064) and Shenzhen Science and Technology Program (No. ZDCY20250901094531003). \textit{(Shibowen Zhang and Jiayang Wu contributed equally to this work). } \textit{(Corresponding author: Yao Su).}}
\thanks{Shibowen Zhang, Jikai Wang, and Zonghai Chen are with Department of Automation, University of Science and Technology of China, Hefei 230027, China. (e-mails: bz24010006@mail.ustc.edu.cn; wangjk@ustc.edu.cn; chenzh@ustc.edu.cn).}
\thanks{Jiayang Wu is with Department of Automation, Tsinghua University, Beijing 100084, China. (e-mails: wu-jy25@mails.tsinghua.edu.cn).}
\thanks{Shibowen Zhang, Jiayang Wu, Guannan Liu, Zhehan Li, Hangxin Liu, Jingwen Zhang, Jiayi Wang, and Yao Su are with State Key Laboratory of General Artificial Intelligence, Beijing Institute for General Artificial Intelligence (BIGAI), Beijing 100080, China (e-mails: liuguannan@bigai.ai; lizhehan@bigai.ai; hxliu@bigai.ai; zhangjingwen@bigai.ai; wangjiayi@bigai.ai; suyao@bigai.ai).}
\thanks{Guannan Liu and Zhehan Li are also with the School of Artificial Intelligence, Xidian University, Xi'an 710126, China.}
\thanks{Helin Zhu, Junjie Liu, Junhong Guo, Xiaokun Leng, and Zhicheng He are with Department of Computer Science, Harbin Institute of Technology, Harbin 150001, China (e-mails: 25S003040@stu.hit.edu.cn; 25S003003@stu.hit.edu.cn; 25S003143@stu.hit.edu.cn; lengxiaokun@hit.edu.cn; hezhicheng@hit.edu.cn).}}
\begin{document}

\maketitle
\begin{abstract}
This paper presents an integrated model-based framework for generating and executing dynamic whole-body dance motions on humanoid robots. The framework operates in two stages: offline motion generation and online motion execution, both leveraging future state prediction to enable robust and dynamic dance motions in real-world environments.
In the offline motion generation stage, human dance demonstrations are captured via a \ac{mocap} system, retargeted to the robot by solving a \ac{qp} problem, and further refined using \ac{to} to ensure dynamic feasibility. 
In the online motion execution stage, a centroidal dynamics-based \ac{mpc} framework tracks the planned motions in real time and proactively adjusts swing foot placement to adapt to real-world disturbances.
We validate our framework on the full-size humanoid robot Kuavo 4Pro, demonstrating the dynamic dance motions both in simulation and in a four-minute live public performance with a team of four robots. Experimental results show that longer prediction horizons improve both motion expressiveness in planning and stability in execution.
\end{abstract}

\section{Introduction}

\IEEEPARstart{H}{umanoid} robots, owing to their anthropomorphic design, have inherent advantages in imitating human movements, making them ideal platforms for entertainment applications such as dance~\cite{kwak2013makes}, which requires expressive whole-body movements to convey emotions and aesthetics~\cite{ravignani2016evolutionary}. Performing dance movements with humanoid robots not only has promising application potential but also represents a step toward enhancing their artistic expressiveness. However, realizing such dynamic dance remains a challenging task, especially when it involves whole-body dynamic motions with simultaneous movements of limbs (e.g., leg strides combined with arm swings). These dynamic movements can significantly change the angular and linear momentum of the robot's body. Such changes are usually high-speed, nonlinear, and coupled, requiring precise coordination of limb trajectories and careful regulation of momentum to maintain balance.

\begin{figure}
    \centering
    \includegraphics[width=\linewidth,trim=0cm 0cm 0.2cm 0cm,clip]{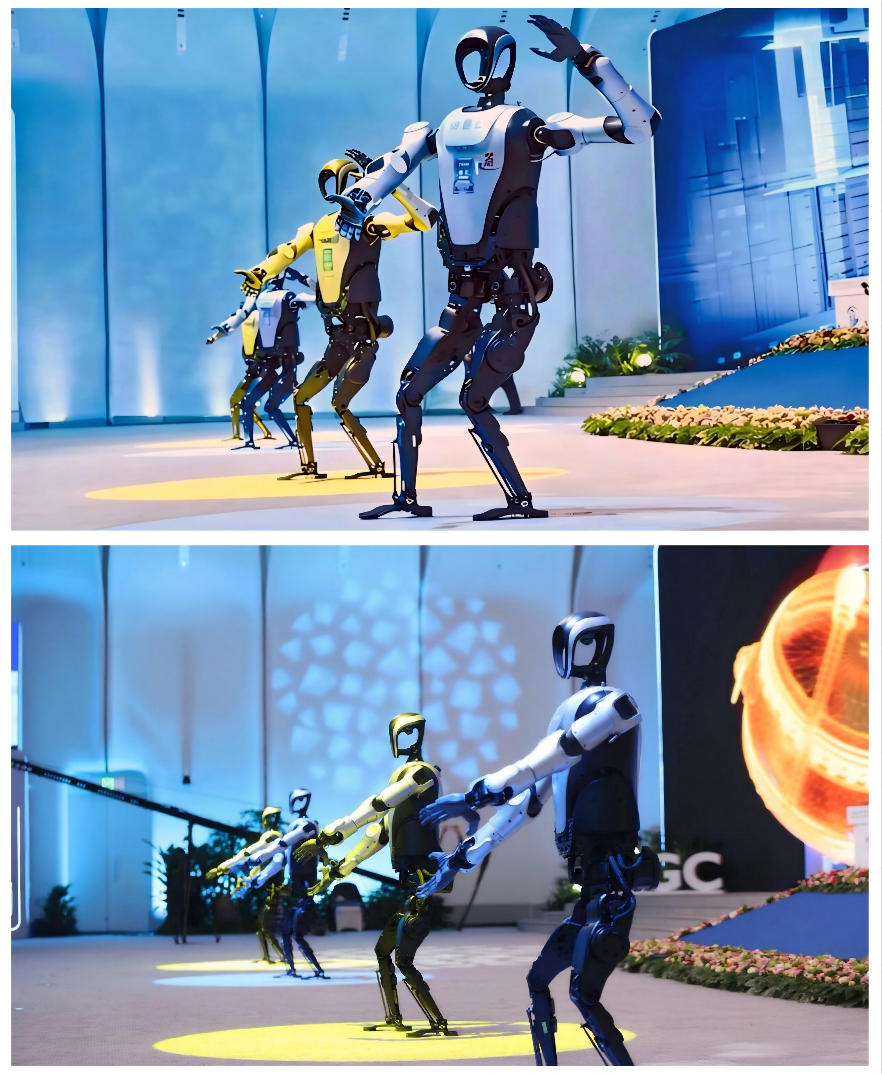}
    \caption{Dynamic whole-body dance motions performed by a team of four humanoid robots.}
    \label{fig:enter-label}
\end{figure}

To achieve this goal, researchers have explored using \ac{rl} to enable humanoid robots to imitate human dance motions captured by a \ac{mocap} system~\cite{he2025asap,cheng2024expressive}, which typically requires tedious design of reward functions and tuning of weighting parameters. Alternatively, Ramos \etal formulate dance motion generation as an optimization problem that ensures the satisfaction of dynamic constraints at each time step~\cite{ramos2015dancing}. However, this method only considers the current state, leading to conservative strategies and reduced motion expressiveness. To address this limitation, \ac{to} serves as a promising approach, as it can anticipate future motion states over a prediction horizon, enabling more expressive motions. Although it has shown successful implementation examples on humanoid robots such as jumping~\cite{he2024cdm} and running~\cite{dai2014whole}, applying it to dance motion generation is challenging, as dance typically involves long-horizon sequences with frequent and aperiodic contact transitions. These factors increase the computational complexity of optimization, making it difficult to generate feasible motions without well-shaped reference trajectories.

In this work, we present a dance motion generation framework for humanoid robots that leverages \ac{to} guided by a well-shaped motion trajectory to produce dynamic whole-body dance motions. First, human dance motion demonstrations are captured via a \ac{mocap} system~\cite{ramos2015dancing}. Subsequently, a geometric motion retargeting module morphs human \ac{mocap} data into robot motions by solving a \ac{qp} problem, where the generated trajectory may violate dynamic constraints but provides a well-shaped reference trajectory for the following \ac{mpc}-based \ac{to} module. Finally, \ac{to} refines the trajectories over a horizon to satisfy dynamic constraints and guide momentum regulation, with longer prediction horizons enabling the generation of dynamically feasible and expressive dance motions.

However, when executing the motion, the robot may encounter tracking errors caused by the inaccurate dynamics model and sensor noise. These errors can lead to deviations between the actual and planned landing positions of the swing foot, which can cause the \ac{zmp} to exceed the support polygon and result in imbalance. To ensure stable execution, it is essential to adjust the trajectories online to adapt to these disturbances. Therefore, we employ an online centroidal dynamics-based \ac{mpc} framework, which exploits the receding horizon to stabilize the robot with state estimation feedback during execution. By selecting an appropriate prediction horizon, the controller can proactively adjust foot placement to maintain balance in real time.

To validate the effectiveness of our approach, we evaluated the framework using dynamic whole-body dance motions characterized by synchronized arm swinging and leg stepping. We deployed the framework on the full-size humanoid robot Kuavo 4Pro, which performed at the opening ceremony of the 2025 Zhongguancun (ZGC) Forum \footnote{\url{https://www.zgcforum.com.cn/}} as shown in \cref{fig:enter-label}. During the ceremony, we achieved the robust operation of a team of four robots and successfully completed a four-minute public dance performance. 

The main contributions of this work are summarized as follows:
\begin{itemize}
    \item We design a \ac{to}-based dynamic whole-body motion generation framework for humanoid robots, ensuring the dynamic feasibility and expressiveness of dance motions.
    \item We employ a centroidal dynamics-based \ac{mpc} framework to achieve robust motion execution.
    \item We successfully deploy the proposed motion generation and control pipeline on a full-size humanoid robot to perform dynamic whole-body dance motions. We further demonstrate the motion on a team of four physical robots in a public performance.
\end{itemize}

We organize the remainder of the paper as follows. We present the relevant literature on humanoid robot dancing in \cref{sec:related work}. Then, we introduce our \ac{to} motion generation framework in \cref{sec:motion_retargeting}. Next, we present the robot control stack for executing dance motions in the real world in \cref{sec:online execution}. The results of simulation and hardware demonstrations are presented in \cref{sec:demonstration}. Finally, we discuss our findings in \cref{sec:discuss} and conclude the paper in \cref{sec:conclusion}.

\section{Related Work}
\label{sec:related work}

\begin{figure*}
    \centering
    \includegraphics[width=1\linewidth,clip]{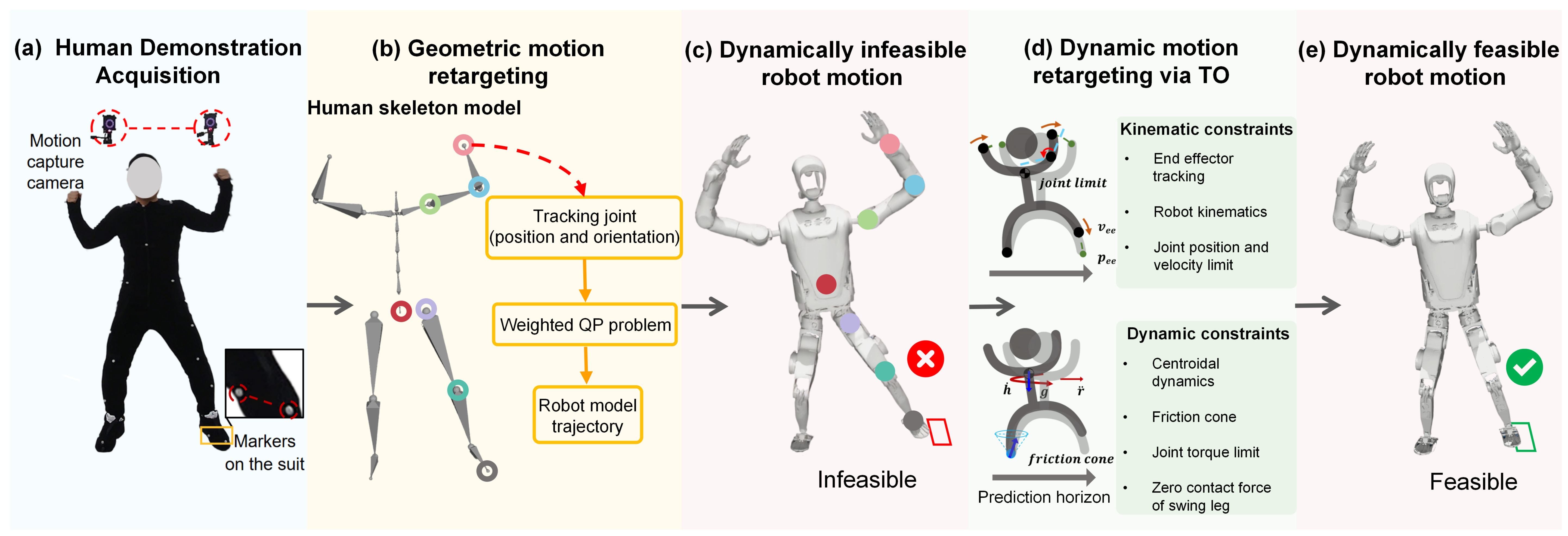}
    \caption{\textbf{Overview of the dance motion generation framework.}
    (a) The \ac{mocap} system captures the motion patterns of markers on the human demonstrator using a high-resolution camera group. (b) Geometric motion retargeting morphs the human motions into robot motions by minimizing the position error between robot joints and human skeleton joints. The corresponding joints of these two models are marked in the same color in (b) and (c). 
    (c) A snapshot of the simulation using the trajectory output by geometric motion retargeting shows a dynamically infeasible motion. 
    (d) Dynamic motion retargeting adapts the trajectory to meet dynamic and kinematic constraints via \ac{to}.
    (e) A snapshot of the simulation using the trajectory after adjustment by dynamic motion retargeting shows a dynamically feasible motion. }
    \label{fig:retarget}
\end{figure*}

Robot dance motions can be composed by blending a set of pre-designed motion primitives. For example, Xia \etal introduced a Markov model to select motion primitives online based on beats and emotions of the music~\cite{xia2012autonomous}.
Similarly, Melya \etal programmed upper-body dance motion sequences for humanoid robots by optimizing selections within a motion library that best match the music features, such as volume and melody~\cite{boukheddimi2022robot}. 
However, these methods focus primarily on upper-body movements, limiting the expressiveness of dance. 

To facilitate expressive whole-body dance movements of humanoid robots that involve simultaneous movement of both upper and lower limbs, a popular approach is to imitate human motions by motion retargeting. The application of \ac{rl} frameworks to imitate human \ac{mocap} data has driven significant progress in generating dance motions. For instance, Cheng \etal utilized \ac{mocap} data to train a whole-body motion generation neural network, which tracks human upper-body motions precisely while relaxing the lower-body imitation to improve stability~\cite{cheng2024expressive}. To improve the dynamics of motions, they further proposed an automated curation method to filter out unsuitable lower-body motion and fine-tuned the pre-training strategy for specific motions~\cite{ji2024exbody2}. However, these approaches often require careful reward tuning, as the effect of adjustments on the resulting behavior is not always straightforward. Moreover, each modification usually entails a lengthy retraining phase, making the process tedious.

Alternatively, researchers have attempted optimization-based methods to imitate human motion.
Kim \etal proposed a motion retargeting method by solving an \ac{ik} problem. This formulation minimized the discrepancy between the captured human motion and the robot motion, while constraining the \ac{zmp} within the support polygon to maintain stability~\cite{kim2009stable}. However,
it only considered \ac{zmp} limitation as the dynamic constraint, which is insufficient to achieve dynamic motions. To address this limitation, Ramos \etal proposed an operational-space inverse dynamics method
within a whole-body \ac{qp} framework. By considering a broader range of dynamic constraints (e.g., joint torque limits), they generated whole-body dance motions~\cite{ramos2015dancing}.
Nevertheless, this approach still struggles to generate motions with large momentum variations, as it only considers the current state. Since momentum accumulation occurs during limb movement, maintaining balance in dynamic motions requires advanced regulation, which relies on the prediction of future states---a capability not supported by instantaneous optimization.

To generate dynamic motion, \ac{mpc} is widely employed by solving a finite horizon \ac{to}~\cite{he2024cdm}. The versatility and dynamics of these motions depend on the selection of the dynamic model. Simple models like the \ac{lipm}~\cite{kajita2003biped} ignore angular momentum and assume constant \ac{com} height, limiting the range of achievable motion. In contrast, \ac{to} based on a whole-body dynamic model can produce highly dynamic motions~\cite{mombaur2009using}, but is computationally intensive. Dai \etal made a tradeoff that combines a centroidal dynamics model and a whole-body kinematics model, generating dynamic motions such as running and jumping. This combination ensures efficient generation of dynamically feasible motions, bridging the gap between the whole-body model and over-simplified models~\cite{dai2014whole}.

However, these methods still face challenges in generating long-duration motions with frequent and aperiodic contact transitions, such as dance, which increases the complexity of the optimization problem. Consequently, directly applying \ac{to} often struggles to produce feasible motion trajectories. To address this, we propose a dance motion generation framework based on the model introduced by Dai \etal~\cite{dai2014whole}. In this framework, we use human \ac{mocap} data and geometric motion retargeting to provide a well-shaped reference trajectory that guides the \ac{to} module. By leveraging the future state predictions, the framework successfully generates dynamically feasible and expressive whole-body dance motions.

\section{Dance Motion Generation}\label{sec:motion_retargeting}

\subsection{Overview}
To achieve dynamic dance motions with humanoid robots, the first step is to design reference motions that are expressive and dynamically feasible. We propose a hierarchical optimization-based motion retargeting framework that maps human demonstrations into feasible whole-body trajectories of the robot offline. As illustrated in \cref{fig:retarget}, the framework comprises three main stages: i) human demonstration acquisition, ii) geometric motion retargeting, and iii) dynamic motion retargeting. In the human demonstration acquisition phase, human motions are captured using an optical \ac{mocap} system. Then, in the geometric motion retargeting stage, we utilize \ac{qp} to adapt human motions into robot joint trajectories. Finally, in the dynamic motion retargeting stage, we refine the trajectories by imposing dynamic feasibility constraints through \ac{to}. The following sections provide detailed descriptions of each stage.

\subsection{Human Demonstration Acquisition}

To acquire human dance demonstrations, a possible approach is to reconstruct human motions from 2D video recordings. However, such methods are sensitive to occlusions and detection errors, which can introduce discontinuities in the reconstructed body movements. In contrast, optical \ac{mocap} systems reconstruct motions by tracking the 3D coordinates of reflective markers attached to the human body, using multiple infrared cameras placed around the demonstrator. This approach can mitigate the impact of occlusions, as the system relies on redundant cameras and markers for reliable tracking and thus achieve robust and high-quality \ac{mocap} data. 

\begin{figure}
    \centering\includegraphics[width=\linewidth,trim=0.1cm 0.3cm 0.1cm 0.1cm,clip]{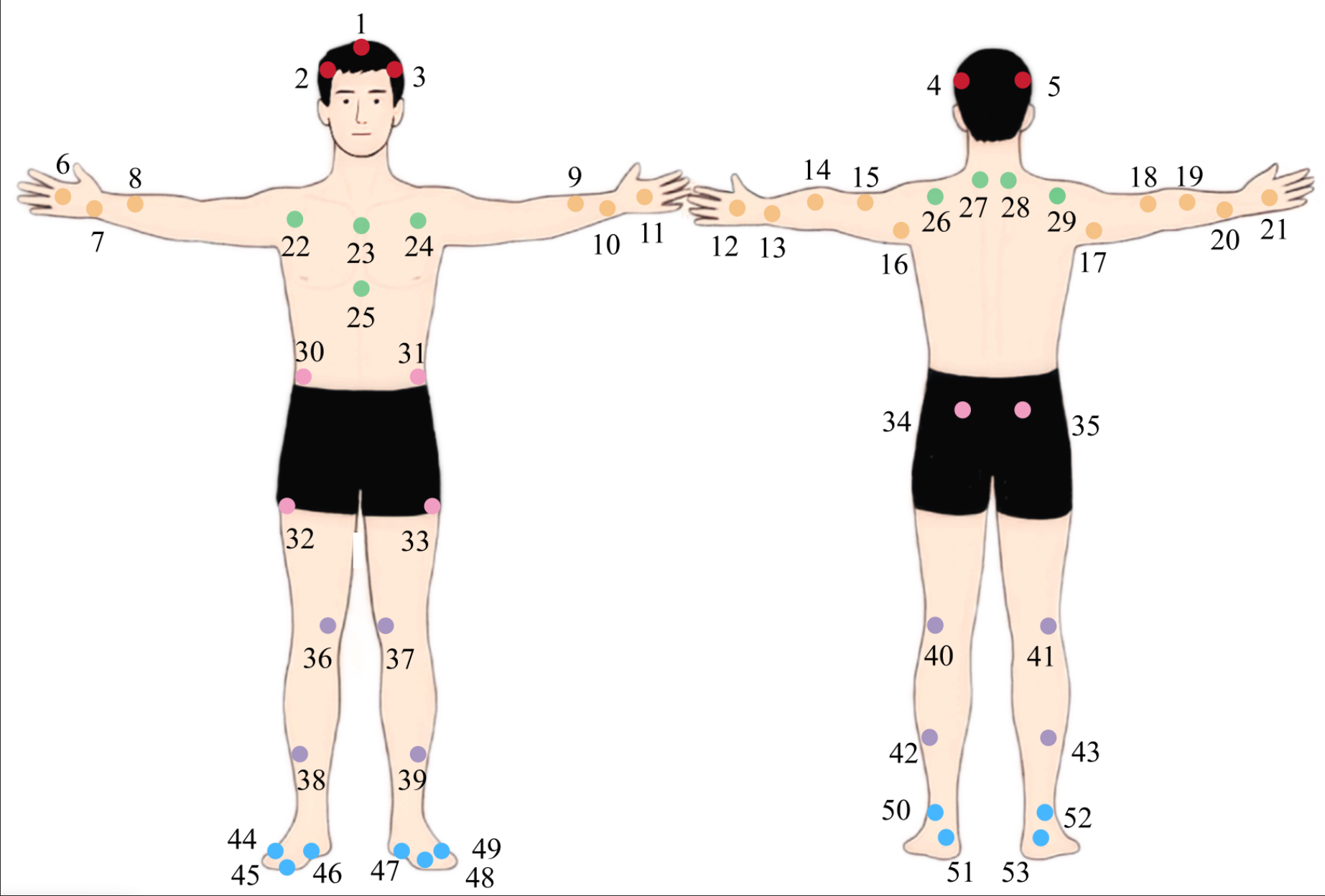}
    \caption{\textbf{The placement of the markers on the human body.} The markers are grouped by different colors, red for head, green for torso, yellow for arms, pink for waist, purple for legs, and blue for feet.}
    \label{fig:MoCap}
\end{figure}

In this work, we use the optical \ac{mocap} system provided by NOKOV\footnote{\url{https://www.nokov.com/}} to record human dance demonstrations. 
This \ac{mocap} system consists of 12 infrared cameras to obtain the spatial trajectories of 53 optical markers attached to the \ac{mocap} suit at 100~Hz. These markers are associated with the body links of a human skeleton model using the provided software. The software groups the markers into clusters, each representing a body segment, and calculates the local coordinate system for each segment. Then, by analyzing the relative transformations between these coordinate systems, the \ac{mocap} system computes the position and orientation of each body link. The placement of the markers is shown in \cref{fig:MoCap}, covering the head (5 markers), torso (8 markers), arms (16 markers), waist (6 markers), legs (8 markers), and feet (10 markers).

\subsection{Geometric Motion Retargeting}

Due to the different body structures between humanoid robots and humans, such as limb length and joint configuration, humanoid robots cannot directly replicate human motion trajectories. For example, humanoid robots typically have shorter legs than humans, and directly executing human leg trajectories in dance motions may result in excessively large strides, compromising balance and causing falls.

To solve this problem, we introduce a geometric motion retargeting method formulated as a \ac{qp} problem for morphing human motions into robot motions.
We first manually establish the joint correspondences as shown in \cref{fig:retarget}(b) and (c) and then proportionally scale the human skeleton to match the robot’s limb lengths. This alignment ensures that the target poses lie closer to the robot’s reachable workspace, thereby improving the kinematic feasibility of the retargeted motions. Subsequently, the \ac{qp} problem minimizes the spatial positional errors between the robot joints and the corresponding scaled human joints.

Specifically, the weighted \ac{qp} formulation is defined as:
\begin{equation}\min_{\Delta{\pmb{q}}}\,\sum_{\mathrm{task}\ i}\left\| \pmb{J}_i\Delta{\pmb{q}}-\pmb{e}_i\right\|_{\pmb{W}}^2 
\label{eq:geometric motion retarget}
\end{equation}

where $\pmb{e}_i=\log_6({\prescript{b_i}{0}{\pmb{T}}}^{-1}\,{\prescript{t_i}{0}{\pmb{T}}})\in se(3)$ is the residual of the pose task. Here, ${\prescript{b_i}{0}{\pmb{T}}}$ and ${\prescript{t_i}{0}{\pmb{T}}}\in SE(3)$ denote the current pose of robot link frame $b_i$ and the target pose of human link frame $t_i$, both expressed in the world frame $0$.
$\pmb{W}$ is the weight matrix. The logarithm $\log_6:SE(3)\to se(3)$ maps the poses into twists. The Jacobian matrix $\pmb{J}_i\in\mathbb{R}^{6\times (6+n_q)}$ is the derivative of task error $\pmb{e}_i$ with respect to joint position $\pmb{q}\in\mathbb{R}^{6+n_q}$, where $n_q$ is the number of robot joints. The joint displacement $\Delta\pmb{{q}}\in\mathbb{R}^{6+n_q}$ is the output of the \ac{qp} solver
and divided by the time step $\Delta t=$ 0.01~s to obtain the commanded velocity. Due to the small time step, the joint-space displacement between consecutive frames remains small, ensuring the validity of the first-order approximation in \cref{eq:geometric motion retarget}. Moreover, by minimizing absolute errors frame-by-frame, this approach prevents approximation errors from accumulating over the trajectory.
 
For the weight matrix $\pmb{W}$ in \cref{eq:geometric motion retarget}, we assign higher weights to the pelvis and feet, which serve as key geometric references for the whole-body configuration. In contrast, the positions of intermediate leg and arm joints are given lower weights, allowing them to accommodate human–robot morphological differences while preserving the demonstrated motion style. This weighting strategy prioritizes the tracking accuracy of pelvis and foot poses, providing a kinematically reasonable geometric reference for the subsequent \ac{to} stage.

Additionally, the \ac{mocap} data only records joint kinematics and lacks contact status between the feet and the ground. We manually annotate contact states for each of the time steps of the captured motion, which are divided into double support, left single support, and right single support. These labels provide a crucial contact schedule for subsequent \ac{to} and robot control algorithms.

\subsection{Dynamic Motion Retargeting via \ac{to}}
The joint trajectories generated by the geometric motion retargeting stage are often dynamically infeasible, as this stage neglects dynamic constraints, such as joint torque limits and contact friction cones. To solve this problem, we introduce a \ac{to} module implemented via \ac{mpc} to enforce these constraints. Given that the dance trajectory contains over $10^4$ frames, directly optimizing the entire sequence would be computationally prohibitive. Therefore, we use \ac{to} in a receding-horizon fashion to iteratively optimize the trajectory, reducing the computation time by breaking the long trajectory into a series of shorter optimization windows. 

Within each window, the optimizer predicts future states using a system dynamics model, solves an optimization problem over the prediction horizon, and applies only the first-step adjustment to the current trajectory segment. Additionally, the receding-horizon mechanism considers dynamic and kinematic constraints at each local optimization step, ensuring the feasibility of the resulting trajectory. In our implementation, processing a complete dance sequence requires approximately 11 minutes on a workstation equipped with an Intel Core i9-13900KF CPU. The details of the \ac{to} module will be described below.

\subsubsection{\textbf{Dynamics model}}
We adopt a centroidal dynamics model, which maps the linear and angular momentum of each link in the multi-body system to the centroidal frame through the \ac{cmm} as:
\begin{equation}
    \pmb{H}=
    \begin{bmatrix}
      M\pmb{\dot{r}}\\ \pmb{h} 
    \end{bmatrix}=
    \pmb{A}(\pmb{q}) \pmb{\dot{q}},
    \label{eq:cmcal}
\end{equation}
where $\pmb{H}\in\mathbb{R}^{6}$ denotes the centroidal momentum vector comprising linear ($M\pmb{\dot{r}}$) and angular ($\pmb{h}\in\mathbb{R}^{3}$) momentum. The system parameters include total mass $M$, \ac{com} position $\pmb{r}\in\mathbb{R}^{3}$, and configuration-dependent \ac{cmm} $\pmb{A}\in\mathbb{R}^{6 \times (6+n_q)}$.

Furthermore, the Newton-Euler equation at \ac{com} of the robot is defined as follows:
\begin{equation}
\begin{bmatrix}
M\pmb{\ddot{r}} \\\dot{\pmb{h}}
 \end{bmatrix}=
\begin{bmatrix}
M\pmb{g}+\sum_{i=1}^{n_c}\pmb{f}_{{c}_{i}} \\
\sum_{i=1}^{n_c}\left((\pmb{{c}}_{i}-\pmb{r})\times \pmb{f}_{{c}_{i}}+\pmb{\tau}_{{c}_i}\right)
\end{bmatrix}
\label{eq:momentrate}
\end{equation}
where $M\pmb{g}$ represents gravitational force, $\pmb{f}_{c_i}, \pmb{\tau}_{c_i} \in \mathbb{R}^3$ denote the contact forces and torques, respectively, and $\pmb{c}_i\in\mathbb{R}^{3}$ specifies locations of the ${n_c}$ active contact points.

Through this model, we can establish a 6D description of the momentum space that incorporates both the linear and angular momentum contributions of robot's limbs. By accounting for limb dynamics and their coupling effects on the whole-body system, this model enables the prediction of dynamic motions.
\subsubsection{\textbf{Optimal Control Problem}}
The optimizer solves an \ac{ocp} over a finite horizon. Inspired by the work of~\cite{sleiman2021unified}, we design the \ac{ocp} function as:
\begin{equation}
\begin{aligned}
 & \min_{\boldsymbol{u}(\cdot)}\,\int_0^TL(\boldsymbol{x}(t),\boldsymbol{u}(t),t)dt \\
 & \mathrm{s.t.~}\,\dot{\boldsymbol{x}}(t)=\boldsymbol{f}(\boldsymbol{x}(t),\boldsymbol{u}(t),t), \\
 & \quad\quad \boldsymbol{g}_n(\boldsymbol{x}(t),\boldsymbol{u}(t),t)=0, \\
 & \quad\quad \boldsymbol{h}_m(\boldsymbol{x}(t),\boldsymbol{u}(t),t)\geq0, 
\end{aligned}
\label{ocp}
\end{equation}
where $L(\pmb{x},\pmb{u}, t)=\left\| \pmb{x}-\pmb{x}^\text{ref}\right\| _{\pmb{Q}}^2+\left\|\pmb{u}\right\|_{\pmb{R}}^2$ is the stage cost, $\|^\cdot\|_{[\cdot]}^2$ denotes the weighted squared $L_{2}$ norm, and $\pmb{Q},\pmb{R}$ are weight matrices. The desired state vector $\pmb{x}^\text{ref}$ is obtained from the retargeted reference trajectory generated in the geometric motion retargeting stage. The state vector $\pmb{x}$ is defined as:
$\pmb{x}=\left[\pmb{\dot{r}}^\mathsf{T},\pmb{h}^\mathsf{T},\pmb{q}^\mathsf{T}\right]^\mathsf{T}\in\mathbb{R}^{12+{n_q}}$. 
The control input vector $\pmb{u}$ is defined as:
$\pmb{u}=\left[\pmb{f}_{c_i}^\mathsf{T},\pmb{v}_j^\mathsf{T}\right]^\mathsf{T}\in\mathbb{R}^{3{n_c}+{n_q}}$, where $\pmb{v}_j$ is the joint velocity.

\begin{figure*}
    \centering
\includegraphics[width=\linewidth]{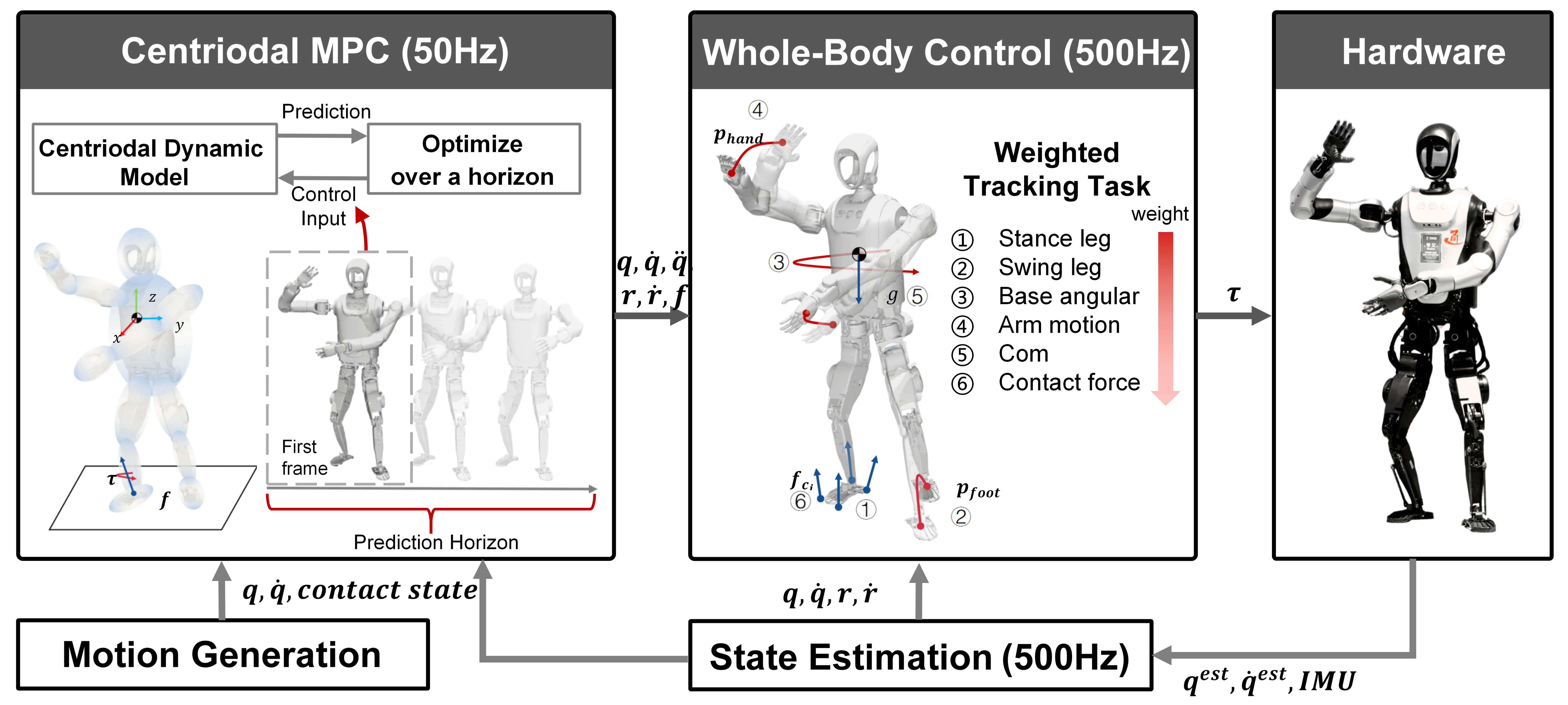}
    \caption{\textbf{Overview of the online motion execution framework.} (i) The centroidal \ac{mpc} takes the offline-generated dance trajectory and the robot's current state as inputs to compute an optimal reference state trajectory. (ii) The whole-body control tracks the reference state trajectory and generates torque control commands for the motors.}
    \label{fig:robot_control_framework}
\end{figure*}

For the state and input weight matrices $\pmb{Q}$ and $\pmb{R}$, the weights are designed to balance stability and motion tracking. In $\pmb{Q}$, we assign the highest weights to the base position and orientation to maintain whole-body stability. The centroidal momentum is also assigned relatively high weights to encourage coordinated whole-body motion. Leg joints are moderately weighted to ensure reliable foot placement, while arm joints are weighted lower to prioritize stability over precise upper-body tracking. 
In $\pmb{R}$, all weights are set lower than the state weights. Joint velocities receive relatively higher weights to promote smooth actuation, while contact forces receive the lowest weight to permit sufficient adjustment of ground reaction forces. This weighting structure provides clear priorities across the state and input terms, making the tuning process more manageable.
In practice, we observed that weights related to the base pose and leg joints are more sensitive and require careful adjustment to maintain stability. In contrast, weights for arm joints and input penalties are less sensitive, tolerating moderate changes without compromising stable execution.

The system dynamics constraints consist of \cref{eq:cmcal,eq:momentrate}, which describe the relationship between limb motion and centroidal momentum. Equality constraints are defined according to the contact state (active or inactive). When the foot comes into contact with the ground, a zero-velocity constraint is imposed to prevent slippage. Conversely, when the foot is off the ground, the end effector tracks the reference trajectory and the contact force is constrained to zero. Inequality constraints are set to ensure that the contact force is within the friction cone and to avoid the joint exceeding its limit.

Previous work~\cite{he2024cdm} detailed the formulation of the \ac{mpc} framework and demonstrated its ability to achieve highly dynamic motions such as jumping. In our implementation, we leverage Pinocchio's centroidal dynamics interface to compute the centroidal velocity and angular momentum, which serve as reference states in the optimization~\cite{carpentier2019pinocchio}. Additionally, we set the maximum optimization iterations to 200, which was found sufficient to ensure dynamic feasibility in dance motions.

\section{Online Motion Execution}
\label{sec:online execution}
In this section, we present the robot control framework for executing the dance motion. 
As illustrated in \cref{fig:robot_control_framework}, the robot control framework consists of three components: i) a centroidal \ac{mpc} that continuously optimizes reference robot state trajectories in an online receding-horizon fashion to track the desired dance motions; ii) an instantaneous \ac{wbc} that computes joint torque commands to track the reference motion trajectories generated by the centroidal \ac{mpc}; and iii) a state estimator that provides real-time feedback of the robot’s state. Next, we describe each module in detail.

\subsection{Centroidal MPC}

During online motion execution, real-world disturbances, such as sensor noise and actuator errors, can introduce tracking errors that may compromise the robot's stability. 
Relying solely on reactive feedback to compensate for these errors is often insufficient, as adaptive actions may lag behind the rapid dynamics of dance motions, leading to instability.
To ensure stable execution, our controller also anticipates future motion states and proactively coordinates limb movements. This foresight allows tracking errors to be compensated for in advance, thereby enhancing the robustness of motion execution. A critical aspect of this proactive strategy is the real-time adjustment of the swing-foot trajectory, as it dynamically reshapes the support polygon and directly regulates the position of the \ac{zmp}.

To this end, we employ \ac{mpc} for online execution, using the offline \ac{to} results as reference trajectories.
The online \ac{mpc} adopts a formulation consistent with the \ac{ocp} defined in \cref{ocp}. This consistency minimizes the discrepancies between planning and execution caused by differing constraints or objective functions. 
During online execution, \ac{mpc} dynamically adjusts the reference trajectory at 50~Hz based on real-time state feedback. 
When future predictions indicate potential instability, the controller proactively adjusts the swing-foot placement to reshape the support polygon and maintain the \ac{zmp} within stable bounds. This adjustment improves the robot’s ability to handle disturbances and ensures dynamic stability throughout execution.
To improve computational efficiency, the solver is warm-started using the solution from the previous time step. Following common real-time \ac{mpc} iteration schemes, we execute only one solver iteration at each control cycle \cite{rit2006}.

After the \ac{mpc} update in each cycle, the module extracts the reference targets required to drive the \ac{wbc} from the updated trajectory. Specifically, the joint position and velocity trajectories, centroidal momentum, contact force, and torso orientation are directly obtained from the state vector $\pmb{x}$ and the control input $\pmb{u}$. The joint acceleration trajectory is computed via numerical differentiation, while the \ac{com} and swing-foot trajectories are derived from the kinematic model. These physical quantities collectively provide the references for the subtasks of the \ac{wbc}.

\begin{table}[ht!]
\centering
\caption{\textbf{Overview of the whole-body control tasks and constraints} }
\label{tab:wbc}
\resizebox{0.9\linewidth}{!}{%
\begin{tabular}{llccc}
\toprule
\textbf{Tasks}  & \textbf{Constraints}  \\
\hline
1) Stance Leg Joint Motion Tracking & 1) Joint Limits    \\
2) Swing Leg Joint Motion Tracking & 2) Friction Cone   \\
3) Base Angular Motion Tracking& 3) Robot Dynamics  \\
4) Arm Joint Motion Tracking &   \\
5) \ac{com} Tracking  & \\
6) Contact Force Tracking & \\
\bottomrule
\end{tabular}}
\end{table}

\subsection{Whole-body Control}
To track the reference whole-body motion trajectory computed by centroidal \ac{mpc}, the robot control framework employs a \ac{wbc} running at 500~Hz. Specifically, the \ac{wbc} is formulated as a weighted \ac{qp} problem that simultaneously optimizes joint accelerations, joint torques, and contact forces~\cite{bellicoso2016perception}. 
To ensure dynamic feasibility, the \ac{wbc} considers constraints including robot dynamics, friction cones and joint limits. The objective function of \ac{wbc} is the summation of multiple weighted tasks. We list all the constraints and cost functions considered in our \ac{wbc} in \cref{tab:wbc}. 
Among these, the motion tracking of the stance-leg joints is given the highest weight to keep the stance foot stationary, thereby reducing the risk of instability caused by contact slip. 

The \ac{wbc} employs a PD control scheme with a feedforward term to generate the joint torque commands:
\begin{equation}
   \pmb{\tau}=\pmb{\tau}_f+\pmb{K}_p(\pmb{q}_j^\text{ref}-\pmb{q}_j)+\pmb{K}_d(\dot{\pmb{q}}_j^\text{ref}-\dot{\pmb{q}}_j)
\end{equation}
where the feedforward torque $\pmb{\tau}_f$ is computed by the weighted \ac{qp}, the reference joint positions $\pmb{q}_j^\text{ref}$ and velocities $\dot{\pmb{q}}_j^\text{ref}\in\mathbb{R}^{n_q}$ are computed by the centroidal \ac{mpc}, while the current joint position $\pmb{q}_j$ and velocity $\dot{\pmb{q}}_j\in\mathbb{R}^{n_q}$ are measured by joint sensors, $\pmb{K}_p$ and $\pmb{K}_d\in\mathbb{R}^{{n_q}}$ are the PD control gains. Since \ac{mpc} operates at a lower frequency than \ac{wbc}, we apply linear interpolation to discrete \ac{mpc} trajectories to align them with the high-frequency execution of \ac{wbc}.

\subsection{State Estimation}

Accurate state estimation is critical for online motion execution, as it closes the control loop by providing real-time feedback to adjust motion commands against real-world disturbances. We implement a state estimation module that fuses multi-source sensor data, similar to~\cite{Bledt2018cheetah}. The robot's orientation is estimated using \ac{imu} data (gyroscope and accelerometer), while its base position and velocity are derived by fusing accelerometer measurements with joint state feedback based on the kinematic model.

\begin{figure}
    \centering
    \includegraphics[width=\linewidth,trim=0.1cm 0.1cm 0.2cm 0.1cm,clip]{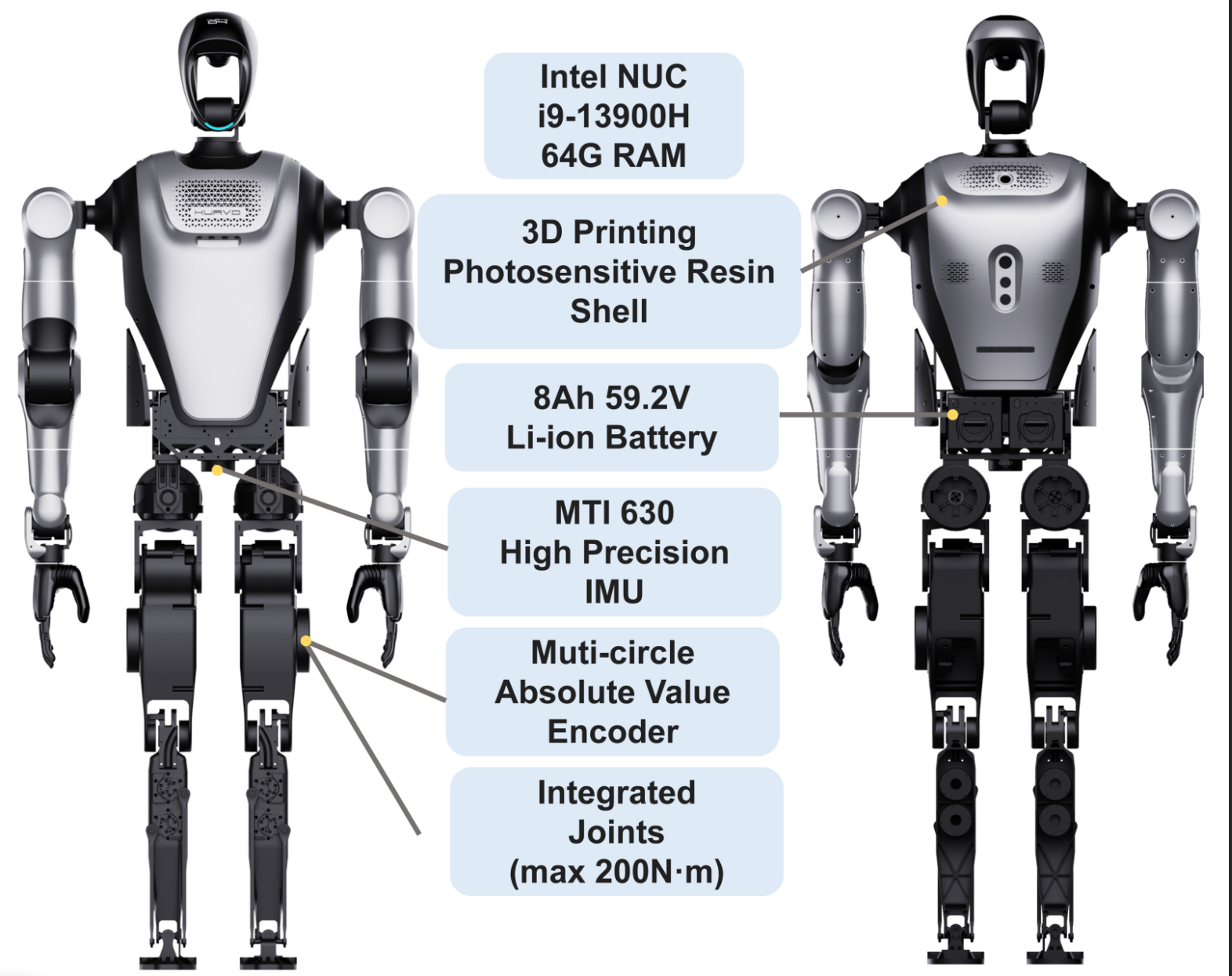}
    \caption{\textbf{Hardware design and configuration of the Kuavo 4Pro humanoid robot.} The total height of the robot is 1.66~m with a mass of 55~kg. Each leg contains 6 \acp{dof} and each arm contains 7 \acp{dof}. The head integrates a 2-\ac{dof} actuation system with a Realsense D435 camera and a Livox Mid360 lidar. The robot contains two computers: an Intel NUC computer (Core i9-13900H) for control and communications, and an NVIDIA Jetson AGX Orin (64~GB) for vision and AI algorithms.}
    \label{fig:hardware}
\end{figure}

\section{Evaluations and Demonstrations}
\label{sec:demonstration}

To validate the effectiveness of our approach, we conducted experiments on the whole-body dance motion in both simulation and the real world using the full-sized humanoid robot Kuavo 4Pro, developed by Leju Robot\footnote{\url{https://www.lejurobot.com/en}}. As shown in \cref{fig:hardware}, this robot has a height of 1.66~m and a weight of 55~kg. Each leg has six \acp{dof}, while each arm has seven \acp{dof}. The large number of \acp{dof} in the limbs enables the generation of expressive whole-body motions that emulate human-like movements. The control framework operates on an onboard computer with an Intel Core i9-13900H CPU and 64~GB RAM. The computer runs a Linux-based operating system with a real-time kernel, ensuring low-latency communication. 

\begin{figure}
\centering
\includegraphics[width=\linewidth,trim=0.1cm 0.0cm 0.0cm 0.0cm,clip]{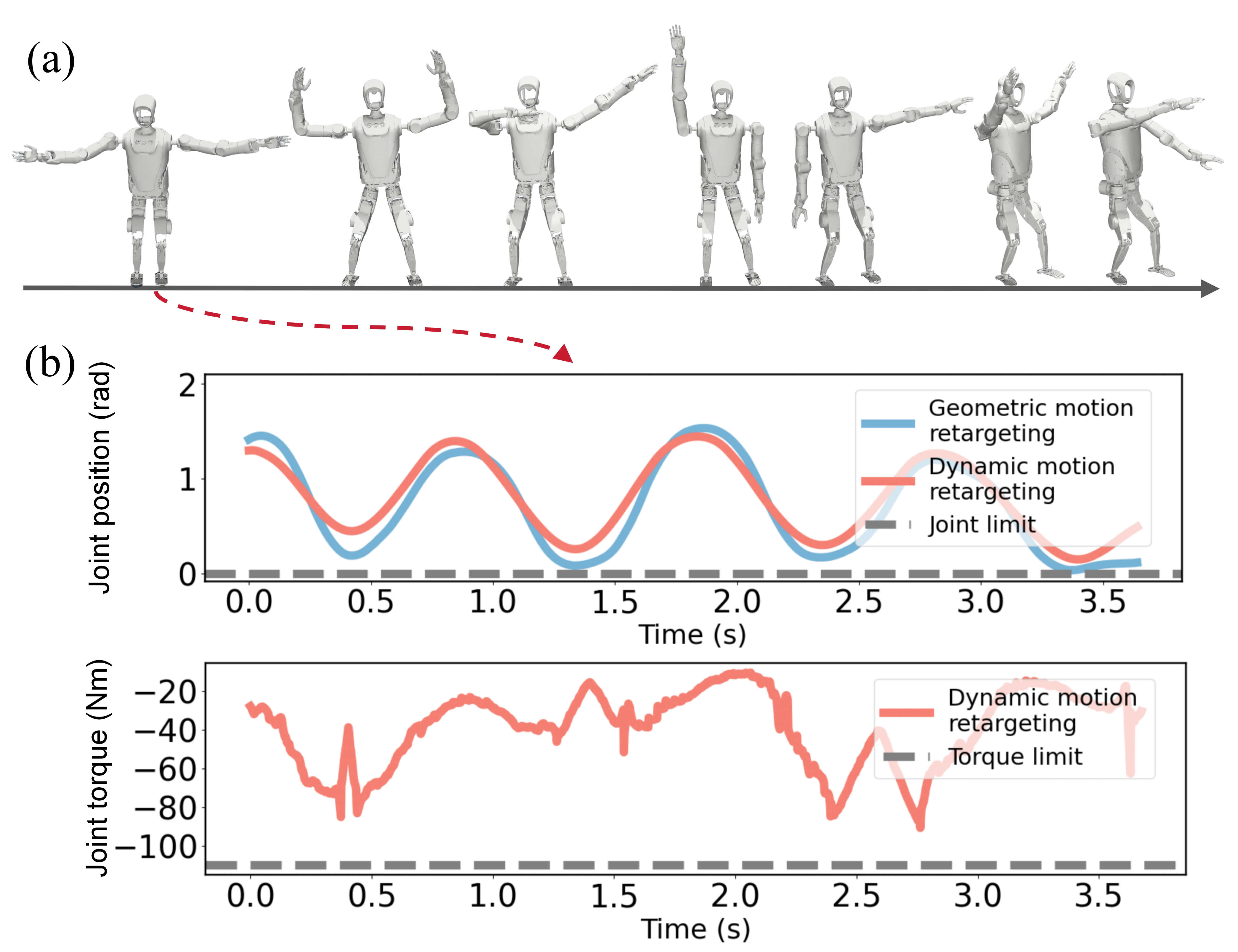}
     \caption{\textbf{Experimental results on enforcing dynamic constraints in the dynamic retargeting module.}  (a) Sequential snapshots of the dance motion in simulation. (b) The top plot shows the knee joint position of the motion pointed by the red arrow. The blue and red lines display the joint position generated by geometric and dynamic motion retargeting, respectively. The gray dashed line shows the knee joint limit. The bottom plot is the knee joint torque of the motion generated by dynamic motion retargeting.}
\label{fig:results}
\end{figure}

To highlight the effectiveness and robustness of our proposed framework, we focus on dynamic motions involving simultaneous movements of both arms and legs (as shown in \cref{fig:results}(a)). Such motions induce large momentum variations, requiring whole-body coordination for dynamic balancing. Our experimental validation was conducted in three stages: i) We evaluated the performance of the dynamic motion retargeting module by comparing its output joint trajectories with those from the geometric retargeting module. The evaluation focused on its capacity to modulate momentum and generate physically feasible trajectories. Furthermore, we investigated how varying prediction horizons influenced the resulting motion momentum. ii) We assessed the controller’s ability to execute motions online under disturbances, focusing on how different prediction horizon lengths influenced foot placement adjustment to maintain balance. iii) We demonstrated the robustness of the entire framework through a four-minute live public performance, where four humanoid robots executed dynamic whole-body dance motions. Next, we describe the experiments in detail.

\subsection*{A. Dynamic Motion Retargeting}
A primary limitation of geometric motion retargeting is its neglect of joint dynamics, which often results in trajectories that are dynamically infeasible due to torque limit violations.
To validate the effectiveness of the proposed dynamic motion retargeting module, we compared its output joint trajectories against those from the geometric motion retargeting module. As shown in \cref{fig:results}(b), our dynamic motion retargeting module explicitly enforced joint torque constraints through \ac{to}, reducing the amplitude of the knee joint motion to keep the torque within limits. This adjustment prevented torque violations and ensured that the generated motion remained dynamically feasible. This highlights the crucial role of considering dynamic constraints in motion generation. 

\begin{figure}[t!]
\centering
\includegraphics[width=\linewidth,trim=0.1cm 0.1cm 0.1cm 0.1cm,clip]{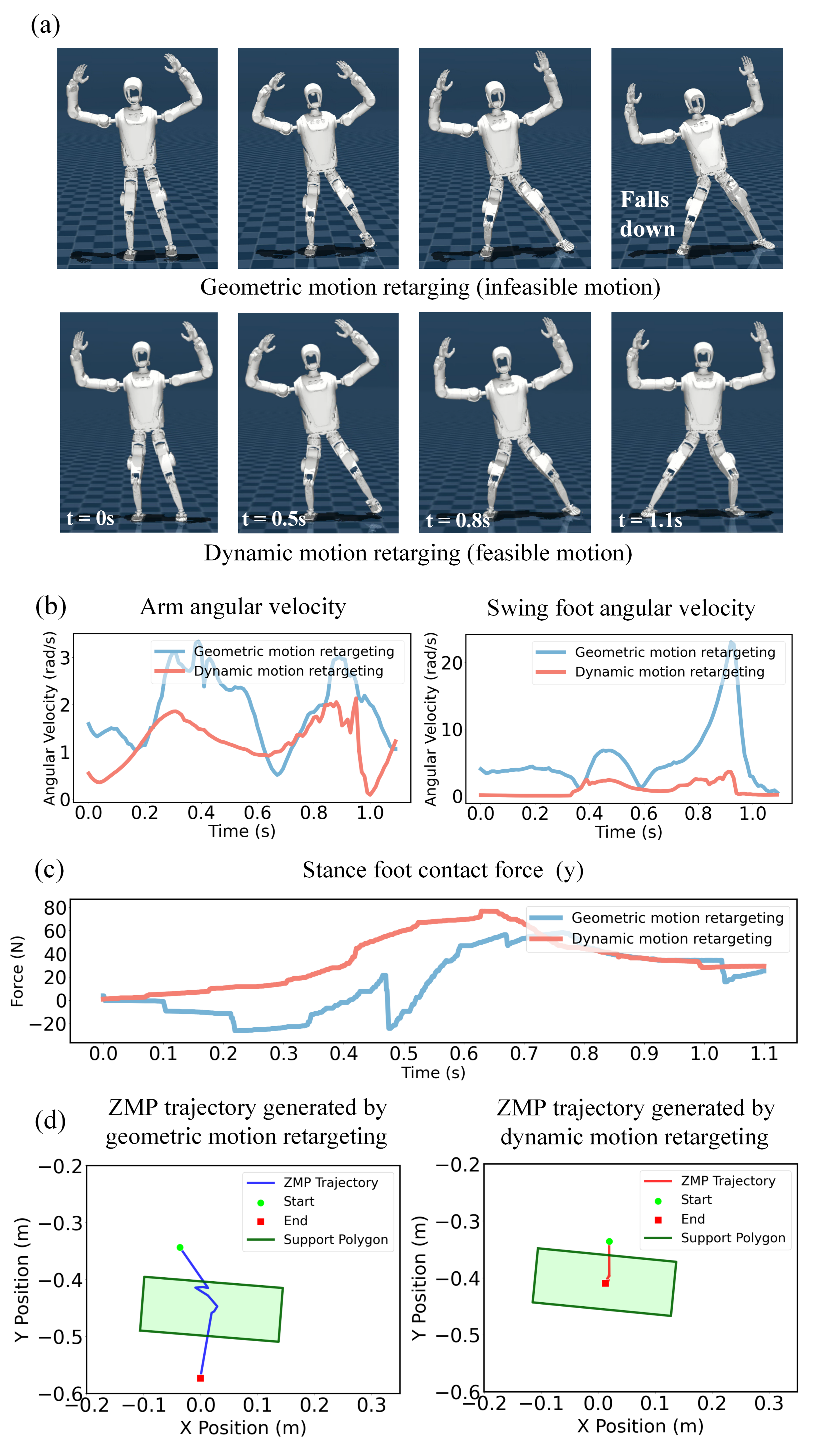}
\caption{\textbf{
Experimental results on evaluating the momentum modulation in the dynamic retargeting module.} (a) Snapshots of the experimental motion sequence. (b) Angular velocities of the arm and swing leg generated by geometric (blue) and dynamic (red) motion retargeting. (c) Stance foot contact force in the y direction of the motion generated by geometric and dynamic motion retargeting. (d) \ac{zmp} trajectories corresponding to geometric (left) and dynamic (right) motion retargeting. The green rectangle indicates the support polygon. }
\label{fig:modulate_momentum}
\end{figure}

In addition to handling dynamic constraints, our module can also modulate momentum by leveraging future states to maintain stability. The snapshot in \cref{fig:modulate_momentum}(a) shows an imbalanced motion during a leg stride coupled with arms swinging. This imbalance occurred because the rapid swing of the leg generated a large angular momentum. To satisfy the conservation of momentum, the upper body needed to produce compensatory momentum, causing the torso to tilt opposite to the direction of the leg motion. When the tilt angle became excessive, the \ac{zmp} exceeded the support polygon boundary, leading to instability. Furthermore, arm swing also contributed to momentum changes, further increasing the risk of imbalance. 

Maintaining balance during such dynamic motions requires momentum modulation through coordinated limb movements. However, optimization methods that rely solely on the instantaneous state lack the ability to anticipate upcoming momentum changes, which potentially result in overly conservative strategies or even unstable motions. By leveraging future states, our module adjusted limb trajectories in advance to facilitate effective momentum modulation. As shown in \cref{fig:modulate_momentum}(b),
the original angular velocity of the swinging leg reached up to 20~rad/s, which was difficult for the torso to compensate for. To maintain balance, our module reduced the angular velocity of the swing leg and arms through optimization over a horizon. Furthermore, as shown in \cref{fig:modulate_momentum}(c), the module generated a larger forward contact force that applied a forward torque to the \ac{com}. This active regulation enabled the robot to maintain a stable posture during large stride movement, ensuring that the \ac{zmp} remained within the support polygon throughout the movement as shown in \cref{fig:modulate_momentum}(d). 

\begin{figure}[t!]
\centering
\includegraphics[width=\linewidth,trim=0.1cm 0.1cm 0.1cm 0.1cm,clip]{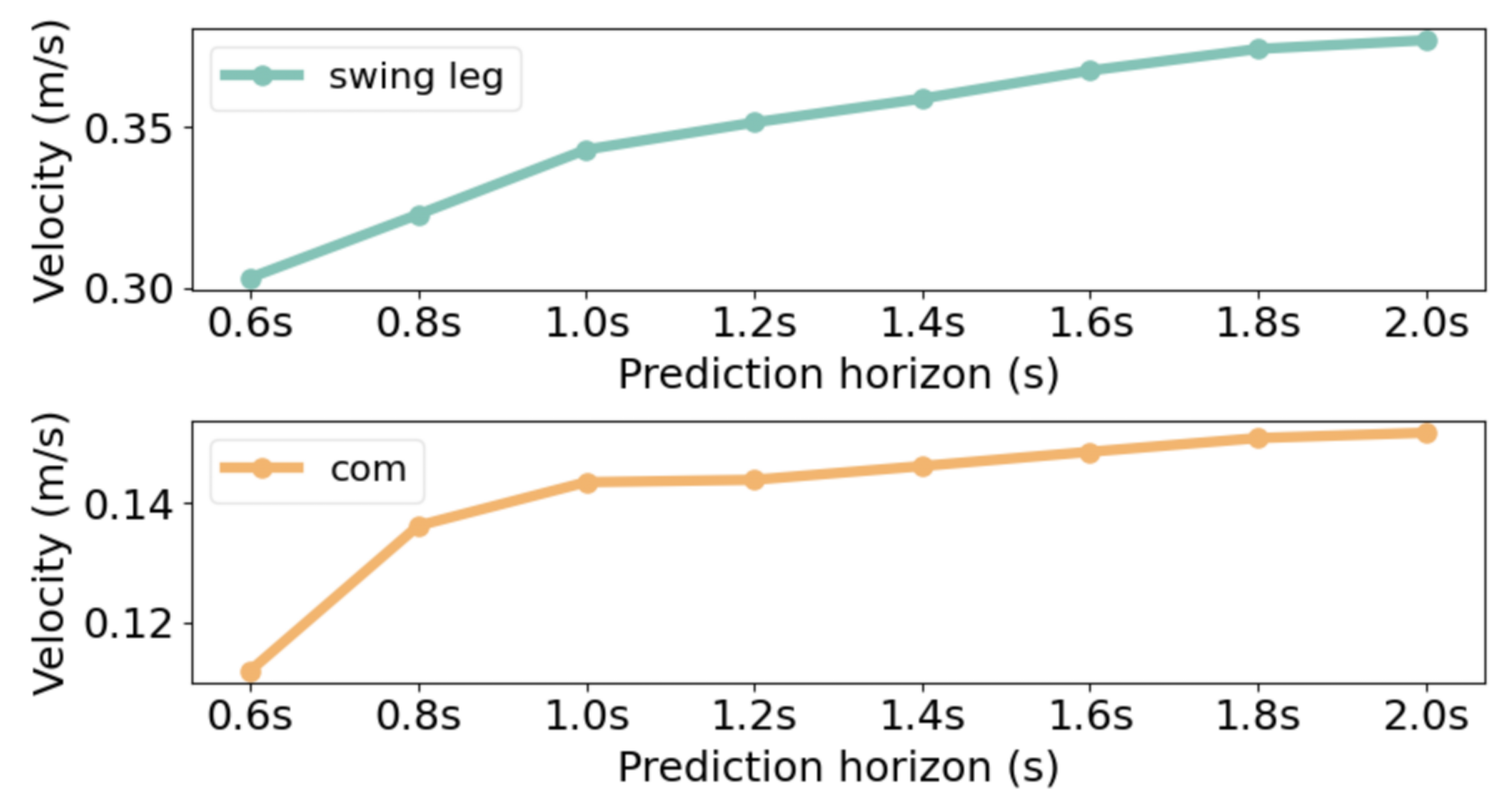}
\caption{\textbf{
Experimental results on evaluating the dynamic capability under different prediction horizons.} (a) Average velocity of the swing leg.  (b) Average velocity of the \ac{com}.}
\label{fig:time_horizon}
\end{figure}

Notably, we observed that the duration of the prediction horizon in \ac{to} affected the dynamic capability of motion, as reflected by the average velocities of the swing leg and the \ac{com}. As shown in \cref{fig:time_horizon}, both velocities increased as the prediction horizon lengthened. 
A substantial increase in velocities occurred between 0.6~s and 1.0~s, after which the increase became more gradual. 
We suspected that shorter horizons resulted in more conservative motions because they failed to cover the full 1.1~s duration of the movement (as shown in \cref{fig:modulate_momentum}(a)). Without anticipating key future events, such as upcoming contact transitions, the optimizer tended to reduce limb speed to ensure stability within the restricted window. While these conservative strategies guaranteed short-term balance, they limited the expressiveness of the resulting motion. In contrast, \ac{to} with longer prediction horizons could predict dynamic changes during the entire motion process. This foresight enabled a trade-off between stride velocity and stability, allowing the robot to generate more expressive movements. The extended horizon further facilitated the proactive coordination of limb momentum and contact forces, making it possible to execute dynamic motions such as large strides while maintaining balance.

\subsection*{B. Online Motion Execution}

\begin{figure}
\centering
\includegraphics[width=\linewidth,trim=0.1cm 0.1cm 0.1cm 0.1cm,clip]{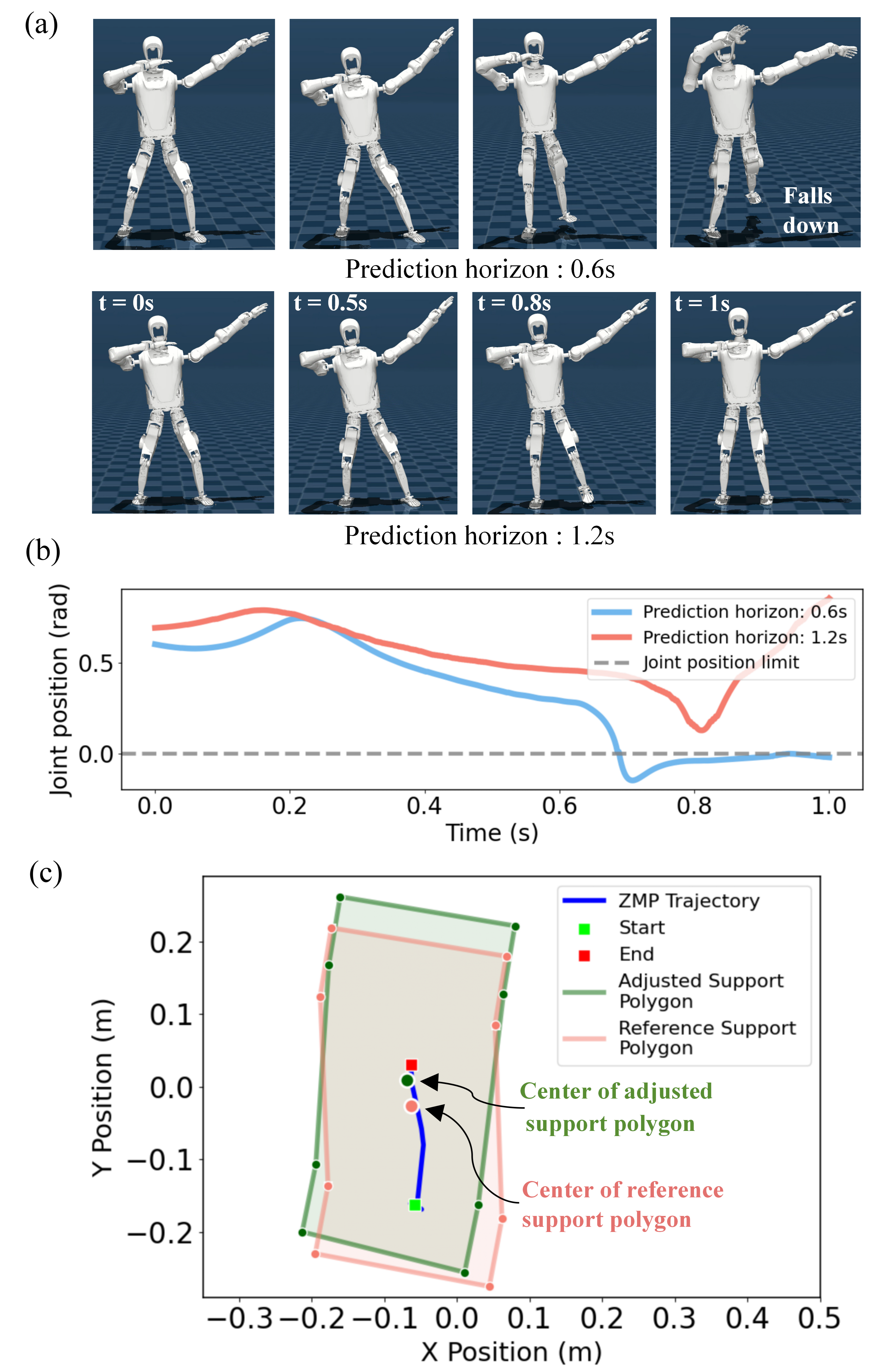 }
\caption{\textbf{Experimental results for assessing the online motion execution module’s ability
to adapt to errors under different prediction horizons.} (a) The top snapshot shows the motion with a prediction horizon of 0.6~s, while the bottom snapshot corresponds to a prediction horizon of 1.2~s. (b) Knee joint trajectories for prediction horizons of 0.6~s (blue) and 1.2~s (red). (c) \ac{zmp} trajectory (blue) and support polygon of adjusted (green) and reference (red) foot placement. The arrow points to the center of the support polygon. }
\label{fig:foot track}
\end{figure}

In real-world environments, robots are often affected by disturbances such as sensor noise and actuator errors, resulting in tracking errors that deviate the robot from its planned trajectory. To maintain dynamic stability under disturbances, the controller must adapt the motion in real time. A critical aspect of this adaptation is foot placement adjustment, which allows the robot to adjust the support polygon in response to tracking errors. This helps maintain the \ac{zmp} within the stable region, preventing imbalance and falls.

The effectiveness of such adaptation depends on the controller’s ability to anticipate future motion, which in our framework is determined by the prediction horizon in \ac{mpc}. 
To evaluate this influence, we conducted experiments with two different prediction horizons: 0.6~s and 1.2~s.
As shown in \cref{fig:foot track}(a), the robot with a short prediction horizon failed to maintain balance, while the 1.2~s horizon enabled successful execution. This failure was caused by the knee joint exceeding its position limit, as shown in \cref{fig:foot track}(b). Specifically, the 0.6~s horizon was unable to cover the entire gait cycle, and future foot placement was not included in the optimization window. As a result, the controller relied heavily on joint motion, particularly the knee joint, to adjust errors. Due to limited foresight, the controller could not anticipate the risk of knee joint limit violations, leading to instability.

In contrast, with the 1.2~s horizon, the controller had sufficient foresight to anticipate upcoming foot contacts and adjust the support region accordingly. As shown in \cref{fig:foot track}(c), the \ac{zmp} trajectory remains centered within the adjusted support polygon rather than the reference support polygon. This indicates that the longer prediction horizon enabled the controller to proactively adjust foot placement, resulting in a support region with a greater stability margin and allowing more effective responses to disturbances.

\begin{table}[ht!]
\centering
\caption{\textbf{Average solve times of \ac{mpc} for different prediction horizons.} }
\label{tab:horizon_time}
\begin{tabular}{ccc}
\toprule
\textbf{Prediction Horizon (s)}  & \textbf{Avg Solve Time (ms)} \\
\hline
0.8 & 14.99  \\
1.0 & 18.06  \\
1.2 & 20.64  \\
1.4 & 24.29  \\
\bottomrule
\end{tabular}
\end{table}
\begin{figure*}[ht!]
    \centering
    \includegraphics[width=\linewidth,trim=0.5cm 0.1cm 0.5cm 0.1cm,clip]{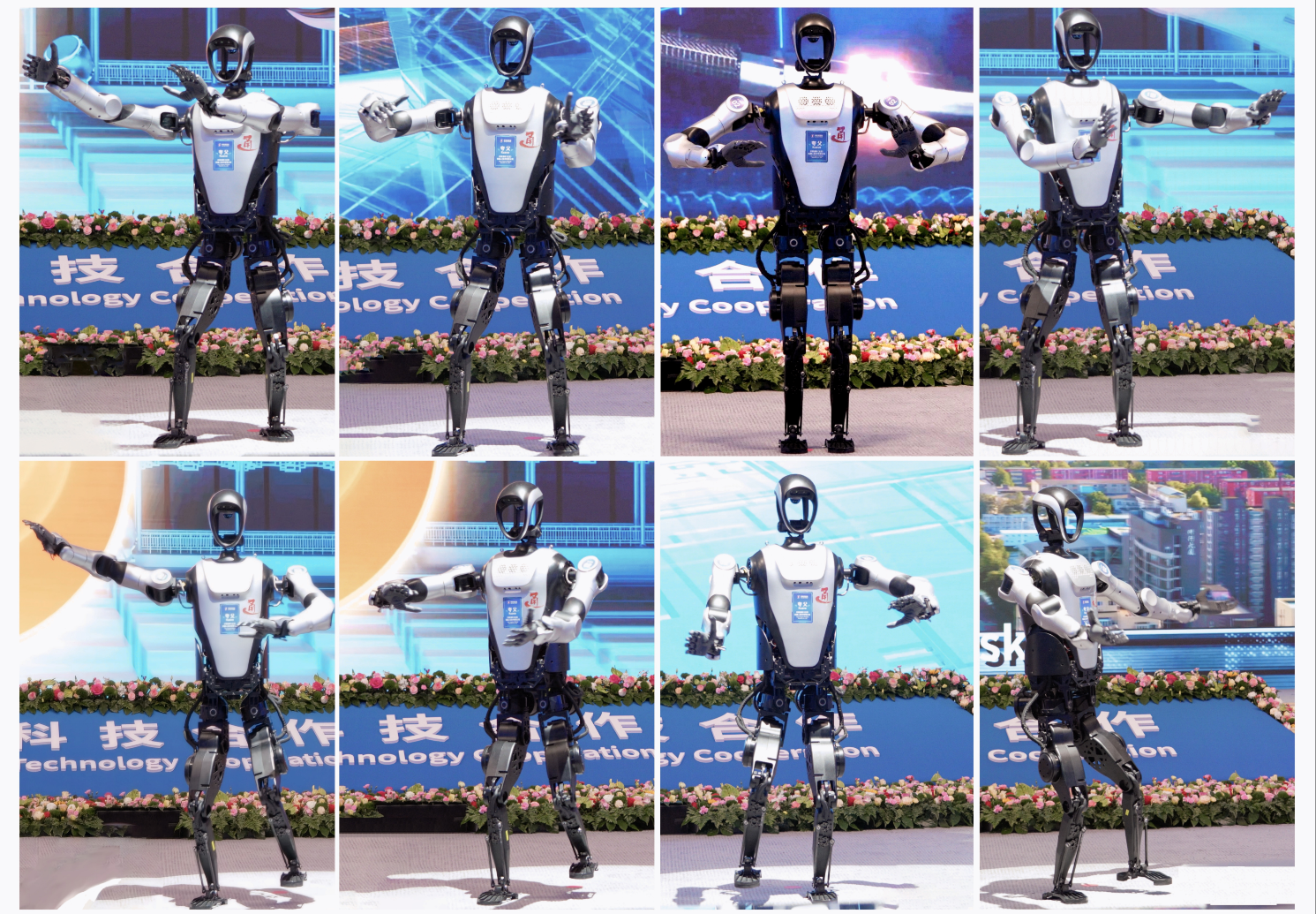}
    \caption{The Kuavo 4Pro humanoid robot performing dynamic whole-body dance motions at the opening ceremony of the 2025 Zhongguancun Forum.}
    \label{fig:performance}
\end{figure*}
While a longer prediction horizon enhances motion stability by enabling better anticipation, it also increases the computational cost of solving the control problem. To ensure that the controller meets real-time execution requirements, we evaluated the average solve time under different prediction horizons, as shown in Table~\ref{tab:horizon_time}. Based on these results, we selected a 1.2~s horizon, which provided sufficient foresight for foot placement adjustment while ensuring that the controller could operate at a frequency close to 50~Hz to meet real-time requirements.

\subsection*{C. Public Performance}

We successfully completed a four-minute dance performance without any failures using a team of four humanoid robots at the opening ceremony of the Zhongguancun Forum, as shown in \cref{fig:performance}. These robots were positioned in a line spaced 3~m apart without safety barriers. The performance faced the following challenges: (1) Dynamic motion complexity: Dynamic whole-body dance motions required precise coordination of limbs to maintain balance; (2) Unstructured environmental disturbances: Unlike rigid laboratory floors, the soft carpets at the venue introduced floor deformation, causing the robots' feet to sink. This altered the expected support polygon and foot landing height, resulting in footstep tracking errors that increased the risk of imbalance.

Despite these challenges, the robots completed the performance through our model-based framework. For motion complexity, the dance motion generation module produced dynamically feasible trajectories that coordinated limb movements and satisfied dynamic constraints through future state prediction. For environmental disturbances, the online motion execution module actively adjusted foot placement to adapt to execution errors in real time via \ac{mpc}.

\section{Discussion}
\label{sec:discuss}
From these experiments, a key insight is the pivotal role of the prediction horizon in both the offline motion generation and online execution phases. Specifically, a longer prediction horizon (1.2~s) enhances motion dynamics during motion generation and improves error adaptation capability during motion execution, resulting in stable and expressive whole-body motions.

In the offline motion generation phase, extending the prediction horizon in the \ac{to} module enables proactive coordination of limb trajectories. This is especially crucial for highly dynamic movements such as leg strides combined with arm swings. In such scenarios, instantaneous or short-horizon optimization tends to either produce conservative motions to ensure constraint satisfaction or lead to unstable motion if the constraints are relaxed. While conservative motions are safe, they compromise the dynamic capability of dance movements. In contrast, our approach allows coordination of momentum and contact forces over the prediction horizon, yielding dynamically feasible trajectories that better reflect the expressiveness of the demonstrated human motion. 

In the online motion execution phase, a longer-horizon \ac{mpc} controller is essential for maintaining stability under real-world disturbances. A key mechanism that enables this robustness is the proactive adjustment of swing-foot placement. Rather than relying solely on joint-level adjustment, which may lead to joint limit violations, the controller adjusts the planned footstep trajectory based on predicted future states. This allows the robot to reconstruct the support polygon in advance, ensuring that the \ac{zmp} remains within the support polygon.
With a short prediction horizon, the controller lacks foresight of upcoming contact phases. As demonstrated in our experiments, using a 0.6~s horizon, the robot was unable to plan footstep placements in advance. As a result, the controller relied heavily on joint motion, which caused the knee joint to exceed its limit and led to a fall. In contrast, with a longer prediction horizon (1.2~s), the controller could foresee potential unstable states and adjust foot placement in advance, thereby guaranteeing stability.

Despite its effectiveness, our framework has several limitations that point to future improvements. First, it requires manual intervention in both reference motion generation and contact schedule annotation. Specifically, although the geometric motion retargeting module provides well-shaped reference trajectories that facilitate \ac{to} convergence, the retargeted trajectories that significantly deviate from dynamic feasibility still require manual adjustment (e.g., shortening excessive stride length). Additionally, the contact schedule requires manual annotation based on the retargeted motion. These dependencies limit the framework's scalability and automation.
Second, the online \ac{mpc} controller incurs a considerable computational cost. The prediction horizon must be carefully selected: longer horizons improve robustness, but increase solve time; shorter horizons reduce cost but weaken foresight. We empirically selected a 1.2~s horizon to maintain a real-time rate of 50~Hz. While this setting was sufficient for the presented dance performances, it may face difficulties in balancing horizon length and solving speed for more highly dynamic movements.

Future work could focus on incorporating learning-based retargeting methods that account for dynamic feasibility to reduce manual adjustment, designing algorithms for automatic contact schedule extraction from raw motion data, and developing a faster \ac{mpc} framework to better handle highly dynamic motions in real time.
\section{Conclusion}
\label{sec:conclusion}
In this work, we presented an integrated model-based framework for dynamic whole-body dance motions on humanoid robots, composed of two key modules: dance motion generation and online motion execution. The dance motion generation module based on \ac{to} leverages well-shaped reference trajectories from the geometric motion retargeting module and optimizes them over a prediction horizon to generate dynamically feasible trajectories by coordinating momentum and satisfying physical constraints. The online motion execution module uses \ac{mpc} to track the generated trajectories in real time, utilizing future state prediction to proactively adjust swing-foot placement and maintain balance under disturbances.

We validated the effectiveness of our framework through simulation and hardware experiments on the full-size humanoid robot Kuavo 4Pro. The results demonstrate that longer prediction horizons enable more expressive motions and more stable execution. Furthermore, we successfully deployed the framework in a four-minute live performance, showcasing its robustness under real-world disturbances.

{
\bibliographystyle{ieeetr}
\bibliography{reference}
}

\end{document}